\documentclass[journal]{IEEEtran}
\usepackage{cite}

\ifCLASSINFOpdf
  \usepackage[pdftex]{graphicx}
\else
\fi
\usepackage{amsmath}
\begin{document}
%
\title{Fisher Discriminative Least Squares Regression for Image Classification}
%
%
%

\author{Zhe~Chen,
        Xiao-Jun~Wu$^*$
        and~Josef~Kittler,~\IEEEmembership{Life~Member,~IEEE}
\IEEEcompsocitemizethanks{\IEEEcompsocthanksitem Zhe Chen and Xiao-Jun Wu are with the School of AI \& CS, Jiangnan University, Wuxi 214122, China. Xiao-Jun~Wu is the corresponding author.
E-mail: jnu\_cz@stu.jiangnan.edu.cn, wu\_xiaojun@jiangnan.edu.cn \protect\\
\IEEEcompsocthanksitem Josef Kittler is with the Centre for Vision, Speech and Signal Processing, University of Surrey, Guildford GU2 7XH, U.K. E-mail: j.kittler@surrey.ac.uk \protect\\ }
\thanks{}}

%
%

\markboth{Journal of \LaTeX\ }%
{}
%



\maketitle

\begin{abstract}
Discriminative least squares regression (DLSR) has been shown to achieve promising performance in multi-class image classification tasks. Its key idea is to force the regression labels of different classes to move in opposite directions by means of the proposed the joint use of the  $\epsilon$-draggings technique, yielding discriminative regression model exhibiting wider margins, and the Fisher criterion. The $\epsilon$-draggings technique ignores an important problem: its non-negative relaxation matrix is dynamically updated in optimization, which means the dragging values can also cause the labels from the same class to be uncorrelated. In order to learn a more powerful discriminative projection, as well as regression labels, we propose a Fisher regularized DLSR (FDLSR) framework by constraining the relaxed labels using the Fisher criterion. On one hand, the Fisher criterion improves the intra-class compactness of the relaxed labels during relaxation learning.  On the other hand, it is expected further to enhance the inter-class separability of $\epsilon$-draggings technique. FDLSR for the first time ever attempts to integrate the Fisher discriminant criterion and $\epsilon$-draggings technique into one unified model because they are absolutely complementary in learning discriminative projection. Extensive experiments on various datasets demonstrate that the proposed FDLSR method achieves performance that is superior to other state-of-the-art classification methods. The Matlab codes of this paper are available at https://github.com/chenzhe207/FDLSR.
\end{abstract}


%
\IEEEpeerreviewmaketitle

\section{Introduction}
%
%
%
%
\IEEEPARstart{L}{east} squares regression (LSR) is a simple yet effective statistical analysis technique that has been widely used in the field of pattern recognition and computer vision. Because it is mathematically tractable and computationally efficient, many variants of LSR with  impact on multi-class image classificaiton have been proposed, such as the weighted LSR \cite{strutz2010data}, partial LSR \cite{wold1984collinearity} and kernel LSR \cite{an2007face}. Besides, ridge regression \cite{cristianini2000introduction}, LASSO regression \cite{tibshirani1996regression}, support vector regression (SVR) \cite{cortes1995support}\cite{van2004benchmarking}\cite{jiao2007fast}\cite{suykens1999least} and logistic regression (LogR) \cite{hosmer2013applied} models are also closely associated with traditional LSR. Instead of learning linear classifiers, some studies extended LSR models to the reproducing kernel Hilbert space to exploit nonlinear correlations between the data \cite{paul2013study}\cite{rahimi2008random}. 

From the classification perspective, it is desirable if the margins between different classes become as large as possible, after transforming the data into the corresponding label space \cite{fan2011local}\cite{shao2013feature}\cite{weinberger2009distance}. However, most of the abovementioned methods assume that the training samples should be projected into a strict binary label space, such as $H$ in (1), which renders the distance between inter-class regression targets  equal to a constant, i.e., $\sqrt{2}$. This is clearly inconsistent with our expectation and fails to accurately reflect the classification ability of a regression model. To solve this problem, Xiang \emph{et al.} \cite{xiang2012discriminative} proposed a discriminative LSR (DLSR) algorithm, which introduces a technique called $\varepsilon$-draggings to encourage the inter-class regression labels moving in the opposite directions, thus effectively enhancing the inter-class separability of regression features. In order to explicitly control the margin of the DLSR model, Wang \emph{et al.} \cite{wang2015msdlsr} proposed a margin scalable discriminative LSR (MSDLSR) model by restricting the number of zeros of dragging values. MSDLSR also proved that DLSR in essence is a relaxed version of $l_2$-norm based support vector machine.  Fang \emph{et al.} \cite{fang2017regularized} proposed a regularized label relaxation algorithm (RLR), which, class-wise, adds a compact graph constraint to the DLSR framework to address the over-fitting problem caused by label relaxation. Based on RLR, Han \emph{et al.} \cite{han2020double} proposed a double relaxed regression (DRR) algorithm to obtain more flexible regression parameters by learning two different projection matrices. Chen \emph{et al.} \cite{chen2020low} proposed a low-rank discriminative least squares regression model (LRDLSR) by class-wisely imposing low-rank constraint on the learned relaxed labels of DLSR.  However, LRDLSR only considers the intra-class similarity of $\epsilon$-dragging labels which is not suitable for the large-scale image classification. Zhang \emph{et al.} \cite{zhang2014retargeted} proposed a retargeted LSR (ReLSR) algorithm to directly learn slack regression targets from data, which performs more accurately in measuring the classification error than DLSR. Wang \emph{et al.} \cite{wang2017groupwise} proposed a group-wise ReLSR model (GReLSR) in which the transition values from the same class of ReLSR are restricted to be similar by utilizing a groupwise constraint. Inspired by the method of label relaxation, Zhang \emph{et al.} \cite{zhang2019guide} constructed a label-guided term to improve the tolerance of the domain adaptation model to label noise.

Except for learning relaxed regression labels, representation learning based classification methods, which can be regarded as LSR based models, have also attracted a great deal of attention. Sparse representation aims at characterising the input sample in the light of a sparse linear combination of the atoms of a given dictionary. Sparse representation based classification (SRC) \cite{wright2008robust} induced sparsity using the $l_1$-norm to constrain the representation vector. The test sample is classified by evaluating which class of samples reconstructs it with the minimum error. However, it is controversial whether the $l_1$-norm is crucial to making SRC work well or not. In fact,  the collaborative representation based classification (CRC) algorithm proposed by Zhang \emph{et al.} \cite{zhang2011sparse}, suggests that it is actually the collaborative mechanism produced by the $l_2$-norm that facilitates effective classification. Subsequently, Cai \emph{et al.} \cite{cai2016probabilistic} proposed a probabilistic CRC (ProCRC), which extended the classification mechanism of CRC into a probabilistic framework. Regardless of the argument between $l_1$-norm and $l_2$-norm,  Xu \emph{et al.} \cite{xu2019sparse} proposed a non-negative representation based classification (NRC) algorithm with just a non-negative constraint on representation, which demonstrated that non-negative representation highlights the contribution of homogeneous samples and simultaneously helps to restrain the representation of heterogeneous samples, thus producing sparse yet discriminative representation. Thereafter, Xu \emph{et al.} \cite{xu2019non} proposed a novel non-negative sparse and collaborative representation (NSCR) classifier based on the observation that simultaneous consideration of non-negativity, sparsity and collaborative mechanism can make the learned representation more discriminative and effective. Unfortunately, if samples are corrupted with large-scale contaminations, the performance of SRC, CRC and NRC may be seriously weakened. To address this problem, low-rank representation (LRR) based algorithms \cite{chen2020noise}\cite{liu2012robust}\cite{liu2011latent} were proposed to recover the clean components from noisy data. Specifically, Liu \emph{et al.} \cite{liu2011latent} proposed a latent low-rank representation (LatLRR) algorithm for recovering from the detrimental effects of missing data. Fang \emph{et al.} \cite{fang2018approximate} proposed an approximate low-rank projection learning method to achieve dimensionality reduction and global optimality of LatLRR \cite{liu2011latent} simultaneously. Besides, Zhang \emph{et al.} \cite{zhang2019manifold} intergrates the low-rank properties into a manifold criterion guided transfer learning (MCTL) for structural consistency between different domains.

In fact, taking the original training samples to represent a test sample directly could not guarantee  that the discriminative information hidden in the data will not inflict high computational burden if the training set is very large. Several methods are based on the idea that learning a dictionary from the original samples can effectively enhance the discriminability of learned sample representation \cite{aharon2006k}\cite{jiang2011learning}\cite{zhang2010discriminative}. Li \emph{et al.} \cite{li2015locality} proposed a  locality constrained and label embedding dictionary learning (LCLE) algorithm which takes the locality and label information of atoms into account jointly. Gu \emph{et al.} \cite{gu2014projective} proposed a projective dictionary pair learning (PDPL) algorithm, which simultaneously learns a synthesis dictionary and a structured analysis dictionary by searching for block-diagonal coding coefficients. Yang \emph{et al.} \cite{yang2014sparse} developed a Fisher discrimination dictionary learning algorithm (FDDL) by imposing the Fisher criterion on the coding coefficients so that the learned representation simultaneously delivers small intra-class  and large inter-class scatter. Based on FDDL, Vu \emph{et al.} \cite{vu2017fast} proposed a low-rank shared dictionary learning algorithm (LRSDL), which exploits a low-rank shared dictionary to preserve common features of the data. Li \emph{et al.} \cite{li2019discriminative} avocated a discriminative Fisher embedding dictionary learning algorithm (DFEDL) that simultaneously constructs Fisher embedding on learned dictionary atoms and representation coefficients. Recently, Sun \emph{et al.} \cite{sun2020discriminative} proposed a discriminative robust adaptive dictionary pair learning algorithm (DRA-DPL)  which uses the $l_{2,1}$-norm to encode the reconstruction error and analysis dictionary simultaneously. DRA-DPL also tries to improve the discriminability of analysis codings using the Fisher criterion.  

Most of above-mentioned dictionary learning algorithms use the time-consuming $l_0/l_1$-norm to regularise the representation coefficients, which means both the training and test times are very long. Besides, although FDDL, LRSDL, DFEDL and DRA-DPL achieve relatively good classification performance, it is not very straightforward because they classify samples by evaluating the class-wise reconstruction error. Thus, in this paper we perform image classification by learning  an efficient linear projection rather than sparse sample representation. Based on the model of DLSR \cite{xiang2012discriminative}, we propose a Fisher discriminative least squares regression algorithm (FDLSR), exploiting the advantages of the $\varepsilon$-draggings technique, and of the Fisher discrimination criterion. 

The main motivations and contributions of our FDLSR algorithm can be summarised as follows:
(1) In DLSR, the non-negative matrix devised for relaxed labels is dynamically updated for all classes jointly. Although the $\varepsilon$-draggings technique increases the distances  of inter-class labels,  the margins between different classes do not change much during iteration, as all the values of the relaxation matrix often are of similar magnitude in each iteration if they are forced to be non-negative. Therefore, our FDLSR uses the Fisher criterion to encourage the margins to widen,  resulting in better  inter-class separability of the learned labels.

(2) The $\varepsilon$-draggings technique causes the labels from the same class to be discrete and uncorrelated. To solve this problem, our FDLSR uses the Fisher criterion to improve the intra-class similarity and compactness of the learned relaxed labels, which is equally crucial to learning the discriminative projection.

(3) FDLSR is the first ever method that incorporates the $\varepsilon$-draggings technique and the Fisher discrimination criterion into a common regression model. These two techniques complement each other  so that the obtained regression labels are not only relaxed but also sufficiently discriminative.

The rest of this paper is organized as follows. Section II overviews the related works. The proposed FDLSR algorithm,  optimization method and classification approach are introduced in Section III. The experimental results are presented in Section IV. Finally, Section V concludes this paper.

\section{Related work}

In this section, we revisit the DLSR and LRDLSR algorithms.

\subsection{Discriminative Least Squares Regression (DLSR)}

Consider a training dataset $X=[X_1,X_2,...,X_c]=[x_1,x_2,...,x_n]\in R^{d\times n}$ from $c$ classes, where $d$ and $n$ denote the sample dimensionality and the number of samples, respectively. $X_i\in R^{d\times n_i}$ denotes the sample matrix of the $i$th class and $n_i$ is the sample number of each class. The common regularization function for LSR model can be formulated as 
$$\min_Q\|QX-H\|_F^2+\beta\|Q\|_F^2, \eqno{(1)}$$
where $Q\in R^{c\times d}$ is the projection matrix, $\beta\geq 0$ is the regularization parameter. $QX$ denotes the features extracted from the original training samples $X$. $H=[h_1,h_2,...,h_n]\in R^{c\times n}$ is the corresponding label matrix which  is defined as:  if sample $x_i$ belongs to class $j$, then the $j$th value in $h_i$ is 1, while the others in $h_i$ are all 0s. 

LSR aims to minimize the least square loss between the extracted features $QX$ and the predefined binary labels $H$. As it is a very simple and convex model, the projection matrix has a closed-form solution. However, the least squares loss used in (1) is unbounded and non monotonous \cite{xiang2012discriminative}\cite{zhang2014retargeted}. Forcing the regression features to approximate strict '0-1' binary labels is not appropriate for exact classification tasks, because the distances between any pair of inter-class regression labels are all equal to $\sqrt{2}$. This evidently conflicts with our expectation that the projected  inter-class features should be as far as possible. To solve this problem, DLSR introduced a technique called $\varepsilon$-draggings to encourage the inter-class margins moving in opposite directions. The model of DLSR can be expressed as
$$\min_{Q,S}\|QX-(H+B\odot S)\|_F^2+\beta\|Q\|_F^2,\ s.t.\ S\geq 0, \eqno{(2)}$$
where $S\geq0$ is the non-negative relaxation matrix that should be updated in the optimization process. $\odot$ denotes the Hadamard-product operator (multiply element-wisely). $B$ is a predefined constant matrix that induces the direction of dragging, and its $i$th row and $j$th column element $B_{ij}$ is defined as
\setcounter{equation}{2}
\begin{eqnarray}
B_{ij} = \left\{
\begin{array}{lcl}
+1, & if & H_{ij}=1, \\
-1, & if & H_{ij}=0, \\  
\end{array}
\right. 
\end{eqnarray}
where "+1" means it points to the positive direction, while "-1" means it points to the negative direction. Here, we take four samples from three different classes as an example to show how the $\varepsilon$-draggings technique relaxes the strict binary label matrix into a slack form. Let
\setcounter{equation}{3}
\begin{equation}       
H=\left[                
  \begin{array}{ccccccccc}   
    1 & 0 & 0 & 0  \\  
    0 & 0 & 1 & 0 \\
    0 & 1 & 0 & 1 \\
  \end{array}
\right]    \in R^{3\times4}       
\end{equation}
denote the corresponding label matrix of these four samples which respectively belong to the first, third, second and third class. We can easily find that the distance between any pair of inter-class labels is a constant $\sqrt{(1-0)^2+(0-1)^2+(0-0)^2}=\sqrt2$. This definition is unable to promote good classification performance of the regression model. After introducing a relaxation term $B\odot S$, we obtain the following new label matrix
\setcounter{equation}{4}
\begin{equation}  
H'=\left[                
  \begin{array}{ccccccccc}   
    1+\epsilon_{11}  & -\epsilon_{12} & -\epsilon_{13} & -\epsilon_{14}  \\  
    -\epsilon_{21} & -\epsilon_{22} & 1+\epsilon_{23} & -\epsilon_{24}  \\
    -\epsilon_{31} & 1+\epsilon_{32} & -\epsilon_{33} & 1+\epsilon_{34} 
  \end{array}
\right]    \in R^{3\times4}         
\end{equation}
where $\{\epsilon_{11},\epsilon_{12},...,\epsilon_{33},\epsilon_{34}\}$ are the non-negative dragging values that constitute the relaxation matrix $S$. From the perspective of improving the inter-class separability of projection, the distance between new inter-class labels is increased to be greater than $\sqrt2$. For example, the distance between the labels of the first and fourth samples is
\setcounter{equation}{5}  
\begin{eqnarray}
\sqrt{(1+\epsilon_{11}+\epsilon_{14})^2+(\epsilon_{21}+\epsilon_{24})^2+(1+\epsilon_{34}+\epsilon_{31})^2}=  \nonumber \\ 
\sqrt{2+(\epsilon_{11}+\epsilon_{14})^2+(\epsilon_{21}+\epsilon_{24})^2+(\epsilon_{34}+\epsilon_{31})^2+} \nonumber \\
2(\epsilon_{11}+\epsilon_{14}+\epsilon_{34}+\epsilon_{31})\nonumber \\
>\sqrt2.
\end{eqnarray}
Besides, from the perspective of exact classification, the class margins of a sample are forced to be greater than 1. For example, the distance between the true class and an incorrect class of the first sample is $1+\epsilon_{11}+\epsilon_{21}>1$ and $1+\epsilon_{11}+\epsilon_{31}>1$. The above two factors are designed to increase the inter-class margins as well as  the distances between regression labels. However, we found  that the class margins do not change much in each iteration and DLSR does not consider the intra-class compactness of the relaxed labels.

\subsection{Low-Rank Discriminative Least Squares Regression (LRDLSR)}
In DLSR, the relaxation matrix $S$ is dynamically updated just with a nonnegative constraint. This means the distances between intra-class regression labels will also be increased, as a result leading to overfitting. For instance, in the original label matrix $H$, the distance between the labels of the second and fourth samples is 0. However, in the relaxed label matrix $H'$, this distance is 
$$dist = \sqrt{(\epsilon_{14}-\epsilon_{12})^2+(\epsilon_{24}-\epsilon_{22})^2+(\epsilon_{32}-\epsilon_{34})^2}. \eqno{(7)}$$
Due to the randomness of $S$, in all likelihood, $dist>0$. That is, the correlation of the labels from the same class is weakened after relaxation, thus the discriminative power of projection matrix will certainly be compromised. Actually, the intra-class similarity of regression labels is equally crucial to learning discriminative projection. To address this problem, based on the model of DLSR, \cite{chen2020low} proposed a low-rank DLSR (LRDLSR) algorithms as follows

\setcounter{equation}{7}  
\begin{eqnarray}
\min_{Q,S,b}\frac{1}{2}\|QX-(H+B\odot S)\|_F^2+\frac{\beta}{2}\|Q\|_F^2+\nonumber \\
\lambda\sum^c_{i=1}\|H_i+B_i\odot M_i\|_*,
\ s.t.\ S\geq 0 
\end{eqnarray}
where $\beta$ and $\lambda$ are trade-off parameters. $\|\bullet\|_*$ denotes the matrix nuclear norm (the sum of singular values). $H_i\in R^{c\times n_i}$, $B_i\in R^{c\times n_i}$, and $M_i\in R^{c\times n_i}$ are the label matrix, constant matrix and relaxation matrix of the $i$th class, respectively. The aim of the third term is to ensure regression features from the same class are compact and similar by class-wisely imposing a low-rank constraint on the $\epsilon$-dragging labels. Under the circumstances, the learned labels of LRDLSR are not only
relaxed but also discriminative, thus leading to more effective projection. Nevertheless, it is very time-consuming to solve the nuclear norm optimization problem because it involves the computation of matrix singular value decomposition. In addition, the inter-class margins of the labels learned by $\epsilon$-dragging are not large enough, provided that the samples present large intra-class variations but very small inter-class differences. 

\section{The proposed FDLSR framework}

As mentioned before, although the $\varepsilon$-draggings technique increases the class margins,  the margins do not typically change much from the first iteration to the end. This is mainly because the dragging values exhibit similar distributions and magnitudes in each iteration, if forced merely by requiring that the relaxation matrix $S$ be non-negative. Consequently, the margins may increase just in the first iteration. To address the above problems, first, we use the Fisher criterion to improve the intra-class compactness and similarity of relaxed labels; second, we use the Fisher criterion based on the $\varepsilon$-draggings to force the class margins to increase during iteration, thus further enhancing the inter-class seperability of the learned projection. Inspired by the Fisher criterion and $\varepsilon$-draggings method, we propose the Fisher discriminative least squares regression (FDLSR) model, which can be formulated as 
\setcounter{equation}{8}  
\begin{eqnarray}
\min_{Q,S}\|QX-(H+B\odot S)\|_F^2+\beta\|Q\|_F^2+ \nonumber \\
\lambda Fisher(H+B\odot S),\ s.t. \ S\geq 0,
\end{eqnarray}
where $Q$ is the projection matrix used for classification, $S$ is the non-negative relaxation matrix that is composed of $c\times n$ dragging values. $H+B\odot S$ denotes the relaxed labels learned by $\varepsilon$-draggings method. As shown in (9), the first term is used to learn discriminative projection $Q$ with relaxed regression labels. The third term aims to use the Fisher criterion to regularize the learned labels. For better understanding and optimization of the $Fisher$ function, we introduce a transition variable $T$ and rewrite our FDLSR model as 
\setcounter{equation}{9}  
\begin{eqnarray}
\min_{Q,S,T}\|QX-T\|_F^2+\alpha\|T-(H+B\odot S)\|_F^2+\beta\|Q\|_F^2 \nonumber \\
+\lambda Fisher(T),\ s.t.\ S\geq 0,
\end{eqnarray}
where $\alpha>0$, $\beta>0$ and $\lambda>0$ are scalars that weighs the corresponding terms in (10). By minimizing the second term, $T$ will approximately be equivalent to the relaxed label matrix $H+B\odot S$. The $Fisher$ function is defined as 
\setcounter{equation}{10}  
\begin{eqnarray}
Fisher(T)=\sum_{i=1}^{c}(\|T_i-M_i\|_F^2-\|M_i-M\|_F^2) \nonumber \\ 
+\|T\|_F^2,
\end{eqnarray}
where $T_i\in R^{c\times n_i}$ denotes the relaxed labels of the $i$th class. $M_i$ consists of $n_i$ identical columns equal to the mean vector of all columns in $T_i$. $M$ includes $n$ identical  columns equal to the mean vector of all columns in $T$. By minimizing the $Fisher$ term, the learned label matrix $T$ will not only be relaxed but also discriminative. Specifically, "relaxed" means that the class margins of the regression labels will be increased and dynamically updated so that the process of learning the projection matrix is more flexible. "Discriminative" means the learned labels will simultaneously promote intra-class similarity and inter-class disparity. Different from DLSR and LRDLSR, the inter-class disparity in FDLSR will be enhanced step by step during iteration. It is worth mentioning that as far as we know, FDLSR is the first-of-its-kind attempt at intergrating the Fisher discrimination criterion and $\varepsilon$-dragging method into a unified optimization framework. 

\subsection{Solution to FDLSR}
According to \cite{fang2017regularized}\cite{xiang2012discriminative}, it is impossible to optimize all variables in (10) simultaneously. Therefore, we update each variable  iteratively, capitalising on their  closed-form solutions in each iteration. In our FDLSR, there are three variables, $Q$, $S$ and $T$, that need to be optimized. They can be updated as follows:

\textbf{Step 1. Update $T$:} Fixing $Q$ and $S$, $T$ can be obtained by minimizing the following problem
\setcounter{equation}{11}  
\begin{eqnarray}
J(T)=\|QX-T\|_F^2+\alpha\|T-(H+B\odot S)\|_F^2+\nonumber \\
\lambda Fisher(T).
\end{eqnarray}
Referring to literature \cite{vu2017fast}, the derivative of $Fisher(T)$ with respect to $T$ is
$$\frac{\partial Fisher(T)}{\partial T}=4T+2M-4\hat M, \eqno{(13)}$$
where $\hat M=[M_1, M_2,...,M_c]$.  $M$ and $\hat M$ are calculated using  $T$ from the last iteration. By setting the derivative  $\frac{\partial J(T)}{\partial T}=0$, we have
\setcounter{equation}{13}  
\begin{flalign}
&T-QX+\alpha T- \alpha(H+B\odot S)+ 2\lambda T+\lambda M-\lambda\hat M=0& \nonumber \\
&\Longrightarrow T = \frac{QX+\alpha(H+B\odot S)-\lambda M+2\lambda\hat M}{1+\alpha+2\lambda}.&
\end{flalign}

\textbf{Step 2. Update $Q$:} Fixing $T$ and $S$, $Q$ can be solved by minimizing the following problem
$$J(Q)=\|QX-T\|_F^2+\beta\|Q\|_F^2. \eqno{(15)}$$
By setting $\frac{\partial J(Q)}{\partial T}=0$, we obtain the closed-form solution of $Q$ as
\setcounter{equation}{15}  
\begin{eqnarray}
&QXX^T-TX^T+\beta Q=0& \nonumber \\
&\Longrightarrow Q=TX^T(XX^T+\beta I)^{-1}.&
\end{eqnarray}

\textbf{Step 3. Update $S$:} Fixing $T$ and $Q$, the optimization problem with respect to $S$ becomes
$$J(S)=\|T-H-B\odot S\|_F^2,\ s.t.\ S\geq 0. \eqno{(17)}$$
In fact, the squared $F$-norm of the matrix can be evaluated element-by-element \cite{xiang2012discriminative}. Thus, minimizing (17) is equivalent to minimizing the following $c\times n$ sub-problems
$$J(S_{ij})=(T_{ij}-H_{ij}-B_{ij}S_{ij})^2,\ s.t.\ S_{ij}\geq 0, \eqno{(18)}$$
where $T_{ij}$, $H_{ij}$ and $B_{ij}$ are the $i$th row and $j$th column element of $T$, $H$ and $B$, respectively. Because $B_{ij}^2=1$, we have 
$$(T_{ij}-H_{ij}-B_{ij}S_{ij})^2=[B_{ij}(T_{ij}-H_{ij})-S_{ij}]^2. \eqno{(19)}$$ 
Considering the non-negative constraint, the optimal solution to $S_{ij}$ is
$$S_{ij}=max(B_{ij}(T_{ij}-H_{ij}), 0). \eqno{(20)}$$

\begin{figure*}[!t]
\centering
\includegraphics[scale=0.31]{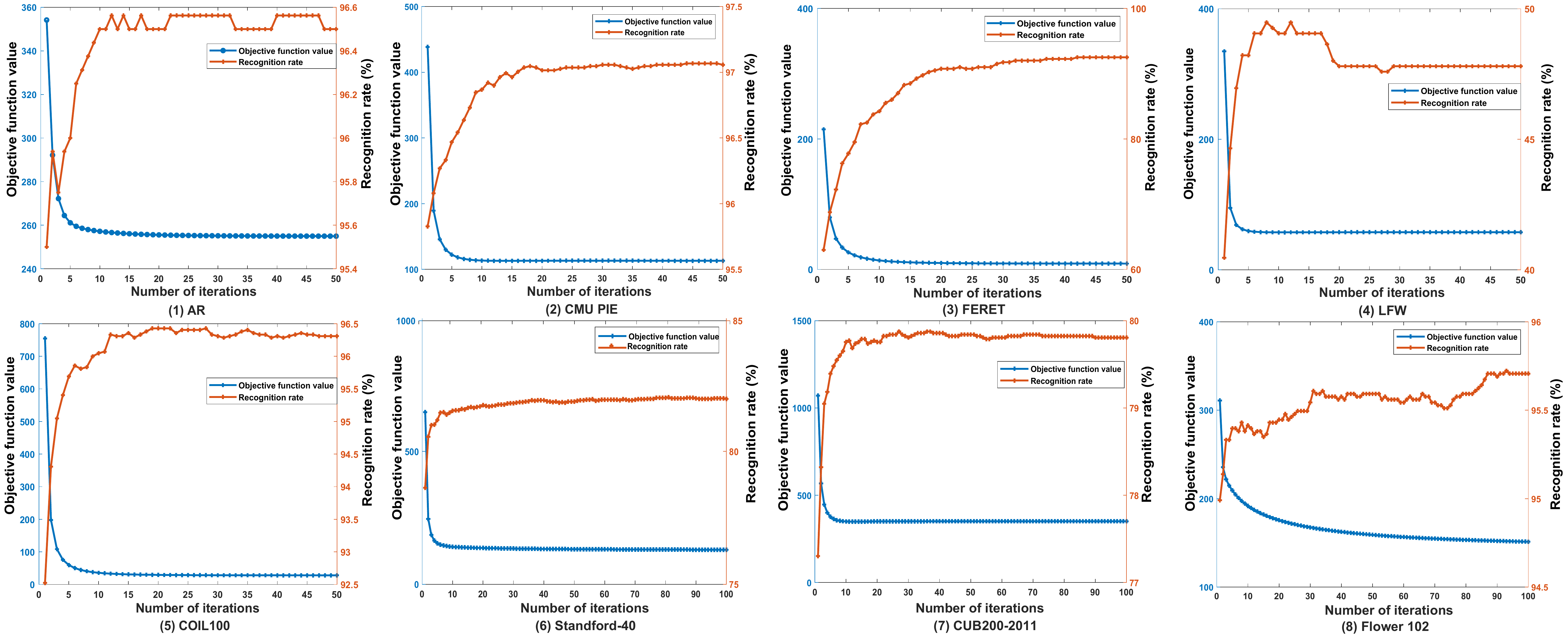}
\caption{The value of objective function and recognition rate versus the number of iterations of the proposed FDLSR algorithm on eight different datasets.}
\end{figure*}

Consequently, the final $S$ can be solved as follows
$$S=max(B\odot (T-H), 0). \eqno{(21)}$$
The optimization steps of FDLSR are summarized in Algorithm 1.

\begin{table}[!ht]
\rule[0.1cm]{8.8cm}{1.5pt}
\leftline {\textbf {Algorithm 1.} \emph{FDLSR}}\\
\rule[0.1cm]{8.8cm}{1.5pt}
\textbf{Input:} Normalized training samples $X$ and their label matrix $H$; Parameters $\alpha, \beta, \lambda$. Maximum
number of iterations $Max\_iter$.

\textbf{Initialization:} $Q=HX^T(XX^T+\beta I)^{-1}$; $S=0^{c\times n}$; $T=H$; $B=2H-1^{c\times n}$.

Let $k=1$, $Q_0=Q$.

\textbf{While} $k<Max\_iter$ \textbf{do}

1. $T = \frac{QX+\alpha(H+B\odot S)-\lambda M+2\lambda\hat M}{1+\alpha+2\lambda}.$

2. $Q=TX^T(XX^T+\beta I)^{-1}.$

3. $S=max(B\odot (T-H), 0). $

4. \textbf{if} $\|Q-Q_0\|_F^2<10^{-4}$, \textbf{then}

  \ \ \ \  \  Stop.

\ \ \ \textbf{end if.}

5. $k=k+1$, $Q_0=Q$.

\textbf{End While}\\
\textbf{Output:} $Q$\\
\rule[0.1cm]{8.8cm}{1.5pt}
\end{table}

\subsection{Algorithm Analysis}

\subsubsection{Complexity analysis} In this section, we analyse the computational complexity of Algorithm 1. 

(1) When updating $T$, the complexity of computing $B\odot S$ is $O(nc)$, the complexity of computing $QX$ is $O(ndc)$. Note that computing $M$ and $\hat M$ does not require much time because both of them just involve calculating one column, so we can neglect their impact. As $d\gg 1$, the final computation complexity of updating $T$ is about $O(ndc)$.

(2) When updating $Q$, the complexity of computing $X^T(XX^T+\beta I)^{-1}$ is $O(nd^2+d^3)$. The complexity of computing $TX^T(XX^T+\beta I)^{-1}$ is $O(ndc)$. Thus, the final computational complexity of updating $S$ is circa $O(ndc+nd^2+d^3)$.

(3) When updating $S$, the complexity of computing $B\odot(T-H)$ is $O(nc)$.

  In many scenarios, the number of training samples and the number of classes are much smaller than the dimensionality of  the feature vector, so the main time-consuming step is one of computing $X^T(XX^T+\beta I)^{-1}$. Fortunately, this term can be pre-computed because its value does not change during iterations. As a result, the final computational complexity of FDLSR is circa $O((nd^2+d^3)+2tndc)$, where $t$ is the number of iterations.

\subsubsection{Convergence analysis and validation}
In this section, we experimentally demonstrate the convergence property of the proposed FDLSR algorithm on eight datasets. The value of the objective function and the recognition rate versus the number of iterations on eight datasets are shown in Fig. 1. It is  apparent that the values of the objective function monotonically decrease to a stable point within a limited number of iterations, which indicates that our algorithm has a good convergence property. Besides, the recognition rate gradually increases during the first several iterations and achieves the saturation within only about 30 iterations. Thus, the computational cost of our method is acceptable. To be fair and for convenience, we report the classification results of FDLSR after 30 iterations on all datasets.

\subsection{Classification}
Once we obtain the optimal projection matrix $\hat Q$, given a test sample $y\in R^d$, we can obtain its projection feature as $\hat Qy$. The nearest-neighbour (NN) classifier is used to obtain the final classification result  by seeking the nearest neighbour of $\hat Qy$ from all the columns in $\hat QX$. This sample is then classified to the class of its nearest neighbor.

\begin{table*}[!t]
\renewcommand{\arraystretch}{1.4}
\caption{Mean Recognition accuracy (mean$\pm$std\%)  and training or test time (s) of different methods on the AR face dataset.}
\label{table_example}
\scriptsize
\centering
\begin{tabular}{|c|c|c|c||c|c|ccc|}
\hline
Train No. & 1 & 2 & 3 & 8 & 11 & 14  & Training time & Test time \\
\hline
SRC \cite{wright2008robust} & 33.99$\pm$1.20 & 59.42$\pm$1.47 & 71.70$\pm$1.13 & 92.55$\pm$0.87 & 95.45$\pm$0.64 & 96.98$\pm$0.39 & None & 19.9285\\

CRC \cite{zhang2011sparse} & \textbf{37.03$\pm$2.06}  & 60.19$\pm$0.72 & 72.36$\pm$0.96 & 91.11$\pm$0.66 & 94.20$\pm$0.78 & 96.10$\pm$0.63 & None & 2.3728\\

ProCRC \cite{cai2016probabilistic} & 35.00$\pm$1.52 & 61.26$\pm$1.31 & 74.44$\pm$1.29 & 92.95$\pm$0.65 & 95.14$\pm$0.46 & 96.83$\pm$0.62& None & 0.3776\\

NRC \cite{xu2019sparse} & 31.84$\pm$1.51 & 58.66$\pm$1.68 & 73.40$\pm$1.35 & 93.92$\pm$0.68 & 96.35$\pm$0.48 & 97.48$\pm$0.39 & None & 160.2309 \\
\hline
LCKSVD2 \cite{jiang2011learning} & 34.55$\pm$1.63 & 55.45$\pm$1.58 & 66.00$\pm$0.94 & 85.72$\pm$1.12 & 90.23$\pm$0.62 & 93.15$\pm$0.56 & 19.4623 & 0.5090 \\

LCLE \cite{li2015locality} & 33.54$\pm$1.28 & 55.50$\pm$1.55 & 69.03$\pm$1.22 & 88.77$\pm$0.85 & 92.80$\pm$0.88 & 94.66$\pm$0.65 & 12.3921 & 0.6557\\

FDDL \cite{yang2014sparse} & 31.84$\pm$1.49 & 57.66$\pm$1.25 & 71.11$\pm$2.16 & 92.40$\pm$0.89 & 94.79$\pm$0.49 & 96.44$\pm$0.31 & 55.0569 & 5.6183\\

LRSDL \cite{vu2017fast} & 32.65$\pm$1.25 & 62.31$\pm$1.28 & 73.59$\pm$0.78 & 92.74$\pm$0.71 & 95.53$\pm$0.46 & 96.84$\pm$0.67 & 181.0698 & 19.9628\\

PDPL \cite{gu2014projective} & 28.14$\pm$1.67 & 52.92$\pm$1.39 & 69.46$\pm$1.39 &  91.88$\pm$0.65 &  95.23$\pm$0.80 &  96.41$\pm$0.54 &  4.6043 & 0.1988 \\

DRA-DPL \cite{sun2020discriminative} & 24.56$\pm$1.55 & 49.79$\pm$1.23 & 65.09$\pm$1.20 &  90.33$\pm$0.50 &  94.06$\pm$0.84 & 95.72$\pm$0.16  & 29.5928 & 0.5988\\
\hline
DLSR \cite{xiang2012discriminative} & 23.61$\pm$1.35 & 47.42$\pm$1.85 & 63.86$\pm$1.65 & 91.62$\pm$0.68 & 95.23$\pm$0.99 & 96.97$\pm$0.50  & 0.5956 & \textbf{0.0257}\\

ReLSR \cite{zhang2014retargeted} & 28.76$\pm$1.81 & 53.75$\pm$1.68 & 69.20$\pm$1.90 & 92.37$\pm$0.54 & 95.61$\pm$0.65 & 97.16$\pm$0.75 & 2.8424 & 0.7750 \\

GReLSR \cite{wang2017groupwise} & 26.68$\pm$1.19 & 52.53$\pm$1.11 & 67.68$\pm$1.04 & 91.78$\pm$0.86 & 94.61$\pm$0.86 & 96.83$\pm$0.49 & 1.9898 & 0.3490\\

DRR  \cite{han2020double} &  22.75$\pm$1.21 & 51.10$\pm$2.23 & 70.35$\pm$0.63 & 93.06$\pm$0.52 & 96.53$\pm$0.41 & 97.37$\pm$0.46 & 4.7895 & 0.0344 \\

LRDLSR  \cite{chen2020low} &  15.02$\pm$2.89 & \textbf{64.84$\pm$1.65} & 79.01$\pm$1.13 & 95.00$\pm$0.70 & 97.08$\pm$0.43 & 98.12$\pm$0.27 & 7.7783 & 0.1008 \\

\hline
FDLSR (ours) & 28.87$\pm$1.65 & 63.99$\pm$1.49  & \textbf{79.41$\pm$1.56} & \textbf{95.32$\pm$0.65} & \textbf{97.25$\pm$0.45} & \textbf{98.18$\pm$0.70} & \textbf{0.1800}  & 0.0358  \\
\hline
\end{tabular}
\end{table*}

\begin{table*}[!h]
\renewcommand{\arraystretch}{1.3}
\caption{Mean Recognition accuracy (mean$\pm$std\%)  and training or test time (s) of different methods on the CMU PIE face dataset.}
\label{table_example}
\centering
\scriptsize
\begin{tabular}{|c|c|c|c||c|c|ccc|}
\hline
Train No. & 1 & 2 & 3 & 10 & 20 & 30 & Training time & Test time \\
\hline
SRC \cite{wright2008robust} & 28.18$\pm$0.97 & 46.45$\pm$1.71 & 59.71$\pm$1.51 & 87.18$\pm$0.40 & 93.94$\pm$0.23 & 95.95$\pm$0.18 & None & 1.2830e+03\\

CRC \cite{zhang2011sparse} & \textbf{29.85$\pm$1.64} & 47.79$\pm$1.33 & 60.14$\pm$1.04 & 86.28$\pm$0.55 & 92.97$\pm$0.40 & 94.78$\pm$0.25 & None & 46.4932 \\

ProCRC \cite{cai2016probabilistic} & 27.66$\pm$1.41 & 48.38$\pm$1.41 & 87.76$\pm$0.54 & 89.11$\pm$0.61 & 94.13$\pm$0.23 & 95.56$\pm$0.20 & None & 4.0425  \\

NRC \cite{xu2019sparse} & 28.10$\pm$1.60 & 46.83$\pm$1.73 & 59.94$\pm$0.77 & 87.98$\pm$0.67 & 94.49$\pm$0.24 & 95.81$\pm$0.19 & None & 3.1190e+03 \\
\hline
LCKSVD2 \cite{jiang2011learning} & 28.76$\pm$1.66 & 42.68$\pm$2.25 & 53.28$\pm$1.76 & 82.99$\pm$0.74 & 91.45$\pm$0.30 & 94.20$\pm$0.22 & 79.8076 & 5.8098 \\

LCLE \cite{li2015locality} & 29.47$\pm$1.17 & 47.14$\pm$0.98 & 58.82$\pm$1.42 & 84.59$\pm$0.74 & 91.88$\pm$0.34& 94.52$\pm$0.19 &  53.6296 & 6.2370 \\

FDDL \cite{yang2014sparse} & 28.67$\pm$1.54 & 47.06$\pm$1.67& 58.13$\pm$1.13 & 81.48$\pm$0.93 & 88.94$\pm$0.50 & 91.24$\pm$0.32 & 120.2367 & 75.7289 \\

LRSDL \cite{vu2017fast} & 28.84$\pm$0.91 & 48.29$\pm$1.17 & 61.49$\pm$0.99 & 87.48$\pm$0.67 & 93.58$\pm$0.45 & 95.46$\pm$0.14 & 1.0185e+03 & 277.5994 \\

PDPL \cite{gu2014projective} & 27.18$\pm$1.61 & 45.29$\pm$1.90 & 58.14$\pm$1.86 & 86.78$\pm$0.68 & 93.19$\pm$0.36 & 95.00$\pm$0.27 & 9.0946 & 2.3520 \\

DRA-DPL \cite{sun2020discriminative} & 17.20$\pm$0.77 & 33.08$\pm$1.11 & 46.11$\pm$2.13 & 82.55$\pm$0.79 & 92.09$\pm$0.37 & 94.91$\pm$0.38 & 69.3391& 9.7962\\
\hline
DLSR \cite{xiang2012discriminative} & 22.56$\pm$1.67 & 43.98$\pm$1.84 & 57.53$\pm$1.47 & 87.67$\pm$0.67 & 93.93$\pm$0.23& 95.78$\pm$0.26 & 1.6614 & 0.2061\\

ReLSR \cite{zhang2014retargeted} & 25.87$\pm$0.93 & 45.38$\pm$2.40 & 58.98$\pm$1.25 & 88.18$\pm$0.92 & 94.22$\pm$0.47 & 96.11$\pm$0.23 & 3.7864 & 6.5152\\

GReLSR \cite{wang2017groupwise} & 27.91$\pm$1.81 & 47,61$\pm$1.96 & 60.41$\pm$1.32 & 87.12$\pm$0.94  & 93.16$\pm$0.42 & 95.23$\pm$0.36 & 2.7138 & 0.3238 \\

DRR  \cite{han2020double} & 23.57$\pm$1.39  & 45.15$\pm$2.01 &  59.26$\pm$0.89 & 87.99$\pm$0.56 & 93.69$\pm$0.35 & 95.90$\pm$0.32 & 19.6434 & 0.2248   \\

LRDLSR  \cite{chen2020low} &  27.35$\pm$0.88 & \textbf{51.26$\pm$1.53} & \textbf{67.10$\pm$2.01} & \textbf{91.57$\pm$0.48} & \textbf{95.78$\pm$0.28} &  \textbf{96.94$\pm$0.14} & 15.6740 & \textbf{0.1008} \\

\hline
FDLSR (ours) & 20.65$\pm$0.65 & 48.56$\pm$2.18 & 63.58$\pm$1.21 & 90.74$\pm$0.69  & 95.47$\pm$0.23 & 96.73$\pm$0.10 & \textbf{0.3570} & 0.2112\\
\hline
\end{tabular}
\end{table*}

\section{Experiments}
In this section, we conduct comparative experiments to verify the effectiveness of the proposed FDLSR algorithm on eight publicly available datasets of three different types: 1) Face datasets: AR \cite{martinez1998ar}, CMU PIE \cite{sim2002cmu}, FERET \cite{phillips2000feret} and LFW \cite{huang2008labeled}; 2) Object datasets: COIL-100 \cite{nene1996object}, CaltechUCSD Birds (CUB200-2011) \cite{wah2011caltech}, Oxford 102 Flowers \cite{nilsback2008automated}; 3) Action dataset: Standford-40 \cite{yao2011human}. We compare our FDLSR with some state-of-the-art algorithms that can be divided into three categories. The first category includes the representation based classification algorithms, namely SRC \cite{wright2008robust}, CRC \cite{zhang2011sparse}, ProCRC \cite{cai2016probabilistic} and NRC \cite{xu2019sparse}. The second category is composed of the dictionary learning algorithms, including the LCKSVD \cite{jiang2011learning}, LCLE \cite{li2015locality}, FDDL \cite{yang2014sparse}, LRSDL \cite{vu2017fast}, PDPL  \cite{gu2014projective} and DRA-DPL \cite{sun2020discriminative}. The number of dictionary atoms are set to the number of training samples. It should be noted that the FDDL, LRSDL and DRA-DPL algorithms use the Fisher discrimination criterion to regularize the representation coefficients of the samples.  The last category includes the least squares regression based classification algorithms, including the DLSR (research baseline of this paper) \cite{xiang2012discriminative}, ReLSR \cite{zhang2014retargeted}, GReLSR \cite{wang2017groupwise} and DRR \cite{han2020double}, and LRDLSR \cite{chen2020low}. We re-implement all the algorithms compared in our study using the source codes provided by the original authors and search the optimal parameters for each algorithm from their original papers. For FDDL, we use the accelerated version of Matlab codes provided by \cite{vu2017fast}. For our FDLSR, the optimal parameters are selected from the candidate set $\{1e-5, 1e-4, 1e-3, 1e-2, 1e-1, 1\}$ by the cross-validation method for seeking the best recognition accuracy.  All the simulations are performed on a PC with Intel (R) Core (TM) i7-8700 CPU @ 3.20 GHz 16G RAM.

\begin{table*}[!t]
\renewcommand{\arraystretch}{1.3}
\caption{Mean Recognition accuracy (mean$\pm$std\%)  and training or test time (s) of different methods on the FERET face dataset.}
\label{table_example}
\scriptsize
\centering
\begin{tabular}{|c|c|c|c||c|ccc|}
\hline
Train No. & 1 & 2 & 3 & 4 & 5 & Training time & Test time \\
\hline
SRC \cite{wright2008robust} & 27.20$\pm$1.34 & 45.27$\pm$1.09 & 55.99$\pm$1.53 & 63.93$\pm$1.58 & 70.35$\pm$1.69 &  None & 64.3648 \\

CRC \cite{zhang2011sparse} & 27.21$\pm$1.02 & 42.00$\pm$1.40 & 50.15$\pm$1.89 & 57.52$\pm$1.66 & 62.18$\pm$1.61 &  None & 1.4012 \\

ProCRC \cite{cai2016probabilistic} & 26.48$\pm$0.76 & 44.81$\pm$1.24 & 57.14$\pm$1.43 & 64.85$\pm$1.84 & 69.50$\pm$1.80 & None & 0.6100 \\

NRC \cite{xu2019sparse} & \textbf{30.16$\pm$0.88}  & 47.85$\pm$1.26 & 59.35$\pm$0.97 & 67.13$\pm$2.13 & 71.22$\pm$2.14 &  None & 28.5290\\
\hline
LCKSVD2 \cite{jiang2011learning} & 24.46$\pm$1,03 & 39.52$\pm$1.17 & 48.59$\pm$1.23 & 55.90$\pm$2.19 & 64.18$\pm$2.19 & 12.8686 & 0.1333\\

LCLE \cite{li2015locality} & 26.98$\pm$1.40 & 41.79$\pm$1.46 & 53.31$\pm$1.21 & 59.23$\pm$1.41 & 62.28$\pm$2.54 & 5.5991 & 0.1944  \\

FDDL \cite{yang2014sparse} & 24.72$\pm$1.22 & 51.04$\pm$1.09 & 61.74$\pm$1.51 & 69.67$\pm$1.68 & 73.93$\pm$1.31 & 74.4847 & 1.8006 \\

LRSDL \cite{vu2017fast} & 27.78$\pm$1.72 & 50.73$\pm$1.06 & 62.46$\pm$1.50 & 70.73$\pm$1.68 & 77.00$\pm$1.67& 231.4636 & 4.7350  \\

PDPL \cite{gu2014projective} & 26.23$\pm$1.28 & 43.78$\pm$2.19 & 54.30$\pm$1.81 & 60.58$\pm$1.65 & 65.30$\pm$2.06 & 33.9260 & 0.4165\\

DRA-DPL \cite{sun2020discriminative} & 22.61$\pm$0.38 & 39.00$\pm$1.01 & 49.49$\pm$1.40  & 55.67$\pm$1.95 & 62.18$\pm$2.38 & 400.7182 & 2.7406 \\
\hline
DLSR \cite{xiang2012discriminative} & 23.26$\pm$1.33 & 46.15$\pm$1.69 & 60.75$\pm$1.71 & 71.45$\pm$1.18 & 79.20$\pm$1.69 &1.6677 & \textbf{0.0145}  \\

ReLSR \cite{zhang2014retargeted} & 23.97$\pm$1.11 & 46.48$\pm$0.81 & 61.78$\pm$1.06 & 73.13$\pm$1.13 & 80.18$\pm$1.26 & 6.4535 &  0.4227 \\

GReLSR \cite{wang2017groupwise} & 29.34$\pm$1.21 & 49.62$\pm$1.17 & 61.91$\pm$1.90 & 70.15$\pm$1.78 & 74.55$\pm$1.04 & 3.1121 & 1.1005\\

DRR  \cite{han2020double} & 19.62$\pm$0.81 & 43.47$\pm$1.26 & 58.85$\pm$1.56 & 70.52$\pm$1.38 & 77.75$\pm$2.07 & 47.5642 & 0.0281\\

LRDLSR  \cite{chen2020low} &  17.08$\pm$1.05 & 63.84$\pm$2.20 & 81.10$\pm$2.25 & 87.68$\pm$0.73 & 90.93$\pm$1.10 &  33.7335 &  0.0933 \\

\hline
FDLSR (ours) & 23.72$\pm$0.94  & \textbf{67.80$\pm$1.10} & \textbf{83.34$\pm$1.44} & \textbf{89.08$\pm$1.01} & \textbf{91.43$\pm$0.91} & \textbf{0.5985} & 0.0238 \\
\hline
\end{tabular}
\end{table*}

\begin{table*}[!h]
\renewcommand{\arraystretch}{1.4}
\caption{Mean Recognition accuracy (mean$\pm$std\%)  and training or test time (s) of different methods on the LFW face dataset.}
\label{table_example}
\scriptsize
\centering
\begin{tabular}{|c|c|c|c||c|c|ccc|}
\hline
Train No. & 1 & 2 & 3 & 5 & 7 & 9  & Training time & Test time \\
\hline
SRC \cite{wright2008robust} & 14.47$\pm$0.97 &  21.71$\pm$0.79 & 26.79$\pm$1.03 & 33.93$\pm$1.53 & 39.24$\pm$1.18 & 44.21$\pm$1.88 & None & 30.5561\\

CRC \cite{zhang2011sparse} & \textbf{16.15$\pm$1.32} & 22.69$\pm$0.99 & 26.97$\pm$1.17 & 34.48$\pm$1.16 & 38.20$\pm$1.12 & 41.70$\pm$1.46 & None & 0.8556\\

ProCRC \cite{cai2016probabilistic} & 15.51$\pm$1.27 & 22.51$\pm$1.21 & 26.78$\pm$1.22 & 33.52$\pm$1.40 & 36.27$\pm$2.43 & 38.99$\pm$2.04 & None & 0.2260\\

NRC \cite{xu2019sparse} & 15.64$\pm$0.91 & 23.32$\pm$1.05 & 29.62$\pm$1.27 & 36.21$\pm$1.58 & 41.99$\pm$1.87 & 45.51$\pm$1.51 & None & 19.3185 \\
\hline
LCKSVD2 \cite{jiang2011learning} & 12.81$\pm$1.10 & 18.81$\pm$1.63 & 23.64$\pm$1.49 & 29.04$\pm$0.79 & 32.27$\pm$1.41 & 35.60$\pm$1.81 & 15.8199 & 0.1256 \\

LCLE \cite{li2015locality} & 16.08$\pm$1.14 & 22.99$\pm$1.35 & 26.51$\pm$1.10  & 33.51$\pm$0.97 & 36.66$\pm$1.42 & 40.04$\pm$1.19 & 2.9857 & 0.1535 \\

FDDL \cite{yang2014sparse} & 16.26$\pm$0.87 & 23.02$\pm$1.01 & 29.12$\pm$1.24 & 36.21$\pm$0.97 & 41.45$\pm$0.94 & 43.40$\pm$1.04 & 8.8339 & 1.0348 \\

LRSDL \cite{vu2017fast} & 16.08$\pm$1.36 & 22.73$\pm$1.38 & 28.25$\pm$1.85 & 37.04$\pm$1.69 & 41.57$\pm$1.03 & 43.84$\pm$2.13 & 71.3349 & 3.5915 \\

PDPL \cite{gu2014projective} & 11.89$\pm$1.27 & 19.93$\pm$1.17 & 24.54$\pm$1.34 & 31.75$\pm$1.35 & 36.10$\pm$1.35  & 41.43$\pm$1.56 & 7.4538 & 0.1415 \\

DRA-DPL \cite{sun2020discriminative} & 10.81$\pm$1.06 & 17.47$\pm$0.95 & 20.85$\pm$1.42 & 27.95$\pm$0.26 & 33.05$\pm$2.39 & 35.28$\pm$1.44  &  68.8842 &  0.6205\\
\hline
DLSR \cite{xiang2012discriminative} & 9.65$\pm$1.07 & 17.52$\pm$1.66 & 22.50$\pm$1.79 & 29.54$\pm$1.64 & 33.81$\pm$1.33 & 38.81$\pm$1.91  & 0.6944 & \textbf{0.0058}\\

ReLSR \cite{zhang2014retargeted} & 11.95$\pm$1.19 & 18.69$\pm$1.94 & 24.54$\pm$1.08 & 31.81$\pm$1.21 & 36.33$\pm$1.26 & 41.51$\pm$1.71 & 1.4460 & 0.1667 \\

GReLSR \cite{wang2017groupwise} & 14.65$\pm$1.41 & 22.54$\pm$1.51 & 27.33$\pm$1.62 & 35.12$\pm$1.18 & 39.98$\pm$1.26 & 43.02$\pm$1.97 & 1.1884 & 0.2308 \\

DRR  \cite{han2020double} & 13.12$\pm$1.41 & 20.77$\pm$1.47 & 25.54$\pm$1.71 & 31.58$\pm$1.17 & 36.89$\pm$2.28 & 39.66$\pm$1.43 & 14.0850 & 0.0124\\

LRDLSR  \cite{chen2020low} &  9.34$\pm$0.88 & 23.57$\pm$1.55 & 29.75$\pm$1.43 & 37.43$\pm$1.21 & 43.24$\pm$1.70 & 45.35$\pm$1.77  & 10.0058 & 0.0782 \\

\hline
FDLSR (ours) & 13.62$\pm$1.11  & \textbf{23.77$\pm$1.63}  & \textbf{30.37$\pm$1.23} & \textbf{37.76$\pm$1.35}  & \textbf{44.10$\pm$1.25} & \textbf{46.18$\pm$1.53} & \textbf{0.2056}  & 0.0166  \\
\hline
\end{tabular}
\end{table*}

\subsection{Experiments for Face and Small Scale Object Recognition}
In this section, we apply FDLSR to four real face recognition datasets and one small-scale object recognition dataset to evaluate the performance of our algorithm.  To be fair, all the algorithms are executed ten times with different random splits of training and test data. 
The average recognition rate, standard deviation of the recognition rates, and the average computing time (including training and test) of different algorithms are reported. In the tables given in the next sub-sections, the symbol $\pm$ denotes the standard deviation of the recognition rates for the ten repetitions. 
To prove whether the proposd FDLSR is able to handle the small-sample-size problem (SSSP) and even the single sample per person (SSPP) problem,  we randomly select 1, 2 and 3 samples per class as the training set on each dataset.

\subsubsection{Experimental results on the AR face dataset} The AR dataset is composed of over 4000 color images of 126 persons (70 men and 56 women) with varying facial expressions, accessory occlusions (sunglasses or scarf) and lighting conditions. Following the setting in \cite{jiang2011learning}, we use a subset in our experiments that only contains expression and illumination variations of 50 men and 50 women. In this subset, each person has 26 samples which are split equally into two sessions. We also use the 540-dimensional feature vector projected by a randomly generated matrix. The parameters are set as $\alpha=1$, $\beta=1e-2$, $\lambda=1$, respectively. We randomly select 1, 2, 3, 8, 11, 14 samples per person for training and the remaining samples are used for testing. The average recognition rates (\%) and computing times (s) of all the algorithms compared are listed in TABLE I.

\subsubsection{Experimental results on the CMU PIE face dataset} The CMU PIE dataset consists of 41368 images of 68 persons, which were taken in different poses, illumination conditions, and facial expressions. Following \cite{chen2020noise}\cite{li2015locality}, we use a subset of PIE which includes five near frontal poses (C05, C07, C09, C27, C29). There are 170 images in total for each person and all images were resized to  32$\times32$ pixels. The parameters are respectively set as $\alpha=1e-1$, $\beta=1e-2$, $\lambda=1e-1$, respectively.   We randomly select 1, 2, 3, 10, 20, 30 images of each person for training and all the remaining samples are used as test samples. The average recognition rates and computing times are shown in TABLE II.

\begin{table*}[!t]
\renewcommand{\arraystretch}{1.3}
\caption{Mean Recognition accuracy (mean$\pm$std\%)  and training or test time (s) of different methods on the COIL-100 object dataset.}
\label{table_example}
\centering
\scriptsize
\begin{tabular}{|c|c|c|c||c|c|ccc|}
\hline
Train No. & 1 & 2 & 3 & 10 & 20 & 30 & Training time & Test time \\
\hline
SRC \cite{wright2008robust} & 48.14$\pm$1.09 & 58.49$\pm$1.39 & 64.81$\pm$0.76 & 83.22$\pm$0.65 & 91.75$\pm$0.54 & 94.74$\pm$0.38 & None & 2.2844e+03\\

CRC \cite{zhang2011sparse} & 46.35$\pm$1.22 & 54.14$\pm$1.12 & 58.16$\pm$0.45 & 75.10$\pm$1.03 & 82.89$\pm$0.83 & 86.45$\pm$0.59 & None & 35.2110\\

ProCRC \cite{cai2016probabilistic} & 42.55$\pm$1.04 & 51.34$\pm$0.66 & 57.02$\pm$0.69 & 73.06$\pm$0.59 & 87.69$\pm$0.41 & 92.61$\pm$0.34 & None & 2.7977 \\

NRC \cite{xu2019sparse} & 44.02$\pm$0.90 & 53.58$\pm$0.90 & 59.54$\pm$1.10 & 75.65$\pm$0.77 & 83.65$\pm$0.53 & 87.53$\pm$0.46 & None & 3.3105e+03\\
\hline
LCKSVD2 \cite{jiang2011learning} & 42.12$\pm$0.78 & 52.25$\pm$0.52 & 58.68$\pm$0.89 & 77.16$\pm$0.88 & 83.48$\pm$1.25 & 81.94$\pm$1.02 & 419.5985 & 4.0134\\

LCLE \cite{li2015locality} & 53.28$\pm$1.10 & 53.29$\pm$0.55 & 58.26$\pm$0.76 & 77.46$\pm$0.59 & 86.56$\pm$0.45 & 91.11$\pm$0.32 & 178.5662 & 5.1868\\

FDDL \cite{yang2014sparse} & 48.65$\pm$1.04 & 58.29$\pm$1.28 & 64.22$\pm$0.55 & 76.62$\pm$0.66 & 81.65$\pm$0.42 & 83.84$\pm$1.04 & 243.3821 & 86.5768 \\

LRSDL \cite{vu2017fast} & \textbf{49.10$\pm$1.44} & 58.48$\pm$0.88 & 64.25$\pm$0.88  & 74.31$\pm$0.66  & 81.98$\pm$0.47 & 84.96$\pm$0.73 & 2.8901e+03 & 223.1466\\

PDPL \cite{gu2014projective} & 45.15$\pm$1.08 & 56.66$\pm$0.91 & 64.18$\pm$0.62 & 83.48$\pm$1.01 & 91.76$\pm$0.39& 94.89$\pm$0.31 & 16.5583 & 1.4496 \\

DRA-DPL \cite{sun2020discriminative} & 42.26$\pm$1.00 & 51.17$\pm$1.58 & 57.54$\pm$1.28 & 74.10$\pm$0.67 & 82.49$\pm$0.51 & 86.36$\pm$0.71 & 111.7775 & 6.5939\\
\hline
DLSR \cite{xiang2012discriminative} & 47.03$\pm$1.03 & 58.91$\pm$1.71 & 67.21$\pm$1.51 & 86.38$\pm$0.27 & 94.01$\pm$0.31 & 96.41$\pm$0.33 & 2.3004  & \textbf{0.1935} \\

ReLSR \cite{zhang2014retargeted} &47.34$\pm$1.17 & 58.96$\pm$1.07 & 67.26$\pm$0.75 & 87.11$\pm$0.76 & 94.16$\pm$0.46 & \textbf{96.61$\pm$0.32} & 8.6083 & 5.7215 \\

GReLSR \cite{wang2017groupwise} & 48.66$\pm$0.91 & 57.94$\pm$0.85 & 62.90$\pm$0.80 & 78.78$\pm$0.51 & 85.70$\pm$0.58 & 89.21$\pm$0.37 & 4.5438 &  0.4627\\

DRR  \cite{han2020double} & 41.72$\pm$0.78 & 52.12$\pm$1.17 & 58.37$\pm$0.47 & 75.50$\pm$0.40 & 91.21$\pm$0.48 & 95.45$\pm$0.10 & 26.9778 & 0.2310 \\

LRDLSR  \cite{chen2020low} & 47.04$\pm$0.89 &  59.01$\pm$1.08 & 67.38$\pm$0.91 & 86.26$\pm$0.88 & 93.66$\pm$0.44 &  96.20$\pm$0.19 & 14.5890 & 0.2903\\

\hline
FDLSR (ours) & 48.97$\pm$0.73 & \textbf{60.44$\pm$1.15} & \textbf{67.62$\pm$0.68} &  \textbf{87.87$\pm$0.43} & \textbf{94.35$\pm$0.53} & 96.48$\pm$0.36 & \textbf{0.5709} & \textbf{0.2010} \\
\hline
\end{tabular}
\end{table*}

\subsubsection{Experimental results on the FERET face dataset} We use a subset of the FERET dataset in which the persons' names are marked with two-letters, i.e., ba, bj, be, bf, bd and bg. This subset includes 1400 face samples from 200 persons and each person has seven different images that were taken in various facial expressions, illuminations, and poses. In the experiments, each image was manually resized to 40$\times$40 pixels. The parameters are respectively set as $\alpha=1e-2$, $\beta=1e-3$, $\lambda=1e-1$.  We randomly select 1, 2, 3, 4, 5 images per person as the training data and the remaining images are used as test data. The experimental results are reported in TABLE III.

\subsubsection{Experimental results on the LFW dataset} The LFW dataset consists of more than 13000 face samples collected from the Internet. It is designed for unconstrained face verification and recognition.  All the images are labeled with the name of the person. Following \cite{chen2020noise}\cite{li2015locality}, we use a subset of LFW which contains 1251 images of 86 persons to conduct test. Each person has about 11-20 images and each image was resized to $32\times32$ pixels. The parameters are set as $\alpha=1e-1$, $\beta=1e-2$, $\lambda=1e-1$, respectively. We randomly select 1, 2, 3, 5, 7, 9 images per person as training samples and treat the remaining images as test samples. The classificaiton accuracies of different algorithms are shown in TABLE IV.

\subsubsection{Experimental results on the COIL-100 dataset}  The COIL-100 dataset consists of 7200 samples from 100 objects that were taken under different lighting conditions and views. Each object has 72 images and  each image was resized to 32$\times$32 pixels in our experiments. The main challenge for classification of this dataset lies in its varying viewpoints. The parameters are respectively set as $\alpha=1e0$, $\beta=1e-1$, $\lambda=1e-1$. We randomly select 1, 2, 3, 10, 20, 30 images per object as training set, and all the remaining samples are used as test set. The classification results are summarized in TABLE V.

In addition, in order to verify whether the Fisher plus $\epsilon$-dragging is more effective for classification than using $\epsilon$-dragging alone, we visualize the extracted features of both training and test samples of DLSR, ReLSR, GReLSR and our FDLSR on the AR dataset using the t-SNE algorithm \cite{boureau2010learning}. The visualization results are shown in Fig. 2.

\begin{figure*}[!h]
\centering
\includegraphics[scale=0.36]{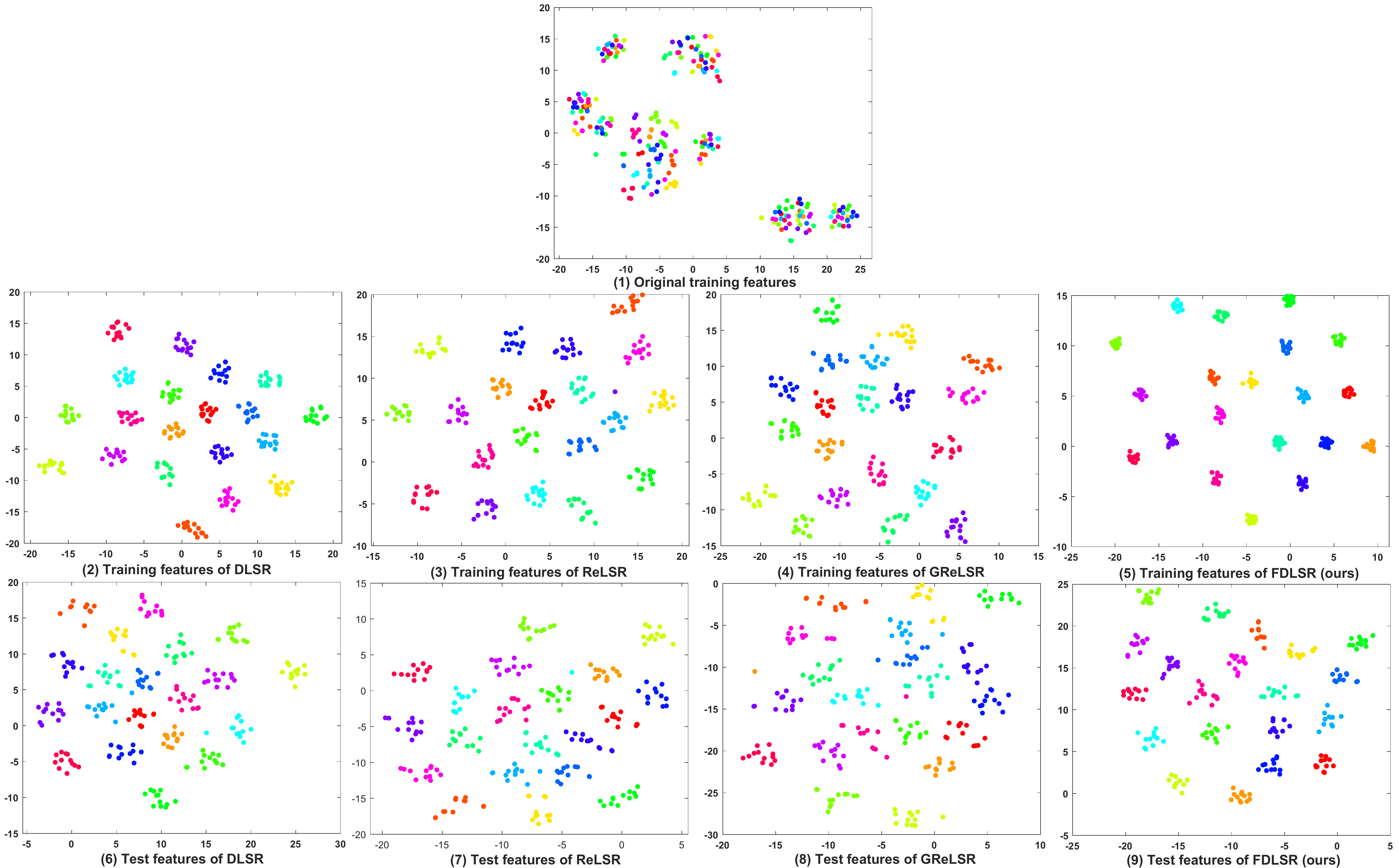}
\caption{T-SNE visualization of features on the AR dataset. The first row: (1) original features $X$; the second row: (2)-(5) extracted training features $\hat Q X$ by DLSR, ReLSR, GReLSR and our FDLSR, respectively; the third row: (6)-(9) extracted test features $\hat Q y$ by different algorithms. In the experiments, random 14 samples per class are selected as the training data and the rest are treated as the test data. Note that we only present the features of the first 20 classes.}
\end{figure*}

\subsection{Analysis of Experimental Results} 

Based on the experimental results shown in tables I-V and Fig. 2, several interesting observations can be made:

(1) \textbf{Overall superiority in recognition performance:} In most cases, our method delivers better recognition performance than the state-of-the-art classification algorithms used for comparison, which shows that our FDLSR can learn discriminative projections for classification. This also demonstrates that by using jointly the Fisher discrimination constraint and the $\epsilon$-dragging technique to build the LSR model, the learned labels will be more relaxed and adequately discriminative than the binary regression labels. Besides, we find that the performance of FDLSR sharply decreases if there is only one training sample per class, but essentially our FDLSR is not designed for the SSPP problem because the Fisher criteron will be ineffective in this condition, as the concept of intra-class similarity ceases to  exist. However, we observe that our FDLSR outperforms its competitors when the number of training samples of each class equals to 2 and 3. This demonstrates that FDLSR has the capability of handling the SSSP problem.

(2) \textbf{Shorter training and test times:} We find that both the training and test time of our FDLSR are lower than those of other the algorithms, specially the representation based classification algorithms, i.e., SRC and NRC, and the dictionary learning algorithms. The main reason is our FDLSR converges very fast within only about 30 iterations and all variables have closed-form solutions.  FDLSR in particular, is more than 100 times faster than the SRC, FDDL, LRSDL algorithms, owing to the fact that all of them use the time-consuming $l_0/l_1$-norm to obtain the sparse coding. Besides, the analysis dictionary learning algorithms, such as PDPL and DRA-DPL, use complementary training matrices to learn the structured analysis dictionary class by class, thus their training phases are less efficient compared to the LSR based algorithms, such as DLSR, ReLSR, GReLSR and our FDLSR.

(3) \textbf{More ideal feature distribution:} From Fig. 2, it is evident that the  features of both the training and test samples extracted by our FDLSR exhibit much better inter-class separability and intra-class similarity than the DLSR and other DLSR based algorithms. Moreover, the inter-class margins achieved by FDLSR are obviously larger than those achieved by DLSR. This proves that the simultaneous use of the Fisher criterion and $\epsilon$-dragging method  helps to improve the compactness of the intra-class regression labels, and thus further increasing the class margins.

(4) \textbf{Comparison between FDLSR and LRDLSR:} The methodologies of LRDLSR \cite{chen2020low} are closely related to the proposed FDLSR. 	From tables I-V, we can find that the classification accuracy of FDLSR is slightly lower than LRDLSR on the CMU PIE dataset, or at least comparable on the AR and LFW datasets. However, FDLSR is significantly better than LRDLSR on the FERET and COIL-100 datasets. This is mainly because the FERET and COIL-100 datasets exist large variations within the same class but small variations across different classes, which is a challenging problem. As thus, it is not enough that LRDLSR only considers the intra-class similarity of $\epsilon$-dragging labels. Different from LRDLSR, FDLSR imposes the Fisher criterion on the relaxed labels, which can further to increase the class separability step by step during iteration, which is good for pattern recognition. Besides, the training process of FDLSR is obviously faster than LRDLSR on all datasets, since all the variables in FDLSR have closed-form solutions and the algorithm converges very fast within only about 10 iterations. Finally, when facing the SSPP problem, FDLSR is superior to LRDLSR on almost all datasets. Because the low-rank constraint in LRDLSR will not work if there is only one training sample per person, that is, LRDLSR degenerates to DLSR. In conclusion, we think LRDLSR and FDLSR are suitable for different scenarios and have different advantages.

\begin{table*}[!t]
\renewcommand{\arraystretch}{1.3}
\caption{Recognition accuracy of different methods on three large-scale image datasets.}
\label{table_example}
\centering
\scriptsize
\begin{tabular}{|c|c|c|c|}
\hline
Algorithms \& Datasets & Standford-40 & CUB200-2011 & Flower 102   \\
\hline
SRC \cite{wright2008robust} (VGG features) & 78.7 & 76.0 & 93.2 \\

CRC \cite{zhang2011sparse} (VGG features)& 78.2 & 76.2 & 93.0 \\

ProCRC \cite{cai2016probabilistic} (VGG features)& 80.9  & 78.3  & 94.8\\

NRC \cite{xu2019sparse} (VGG features)& 81.9 & 79.1 & 95.3\\
\hline
AlexNet \cite{krizhevsky2012imagenet} & 68.6 & 52.2 & 90.4\\

VGG19 \cite{simonyan2014very} & 77.2 & 71.9 & 93.1\\
\hline
DLSR \cite{xiang2012discriminative} (VGG features)& 80.3 & 77.3 & 95.5\\

ReLSR \cite{zhang2014retargeted} (VGG features)& 80.9 & 78.6 &  \textbf{95.8}\\

GReLSR \cite{wang2017groupwise} (VGG features)& 81.3 & 78.8 &  95.5\\

DRR \cite{wang2017groupwise} (VGG features)& 73.4 & 79.0 & 95.8 \\

\hline
FDLSR (ours) (VGG features)& \textbf{82.0} & \textbf{79.8}  &  \textbf{95.8}\\
\hline
\end{tabular}
\end{table*}

\begin{figure*}[!h]
\centering
\includegraphics[scale=0.37]{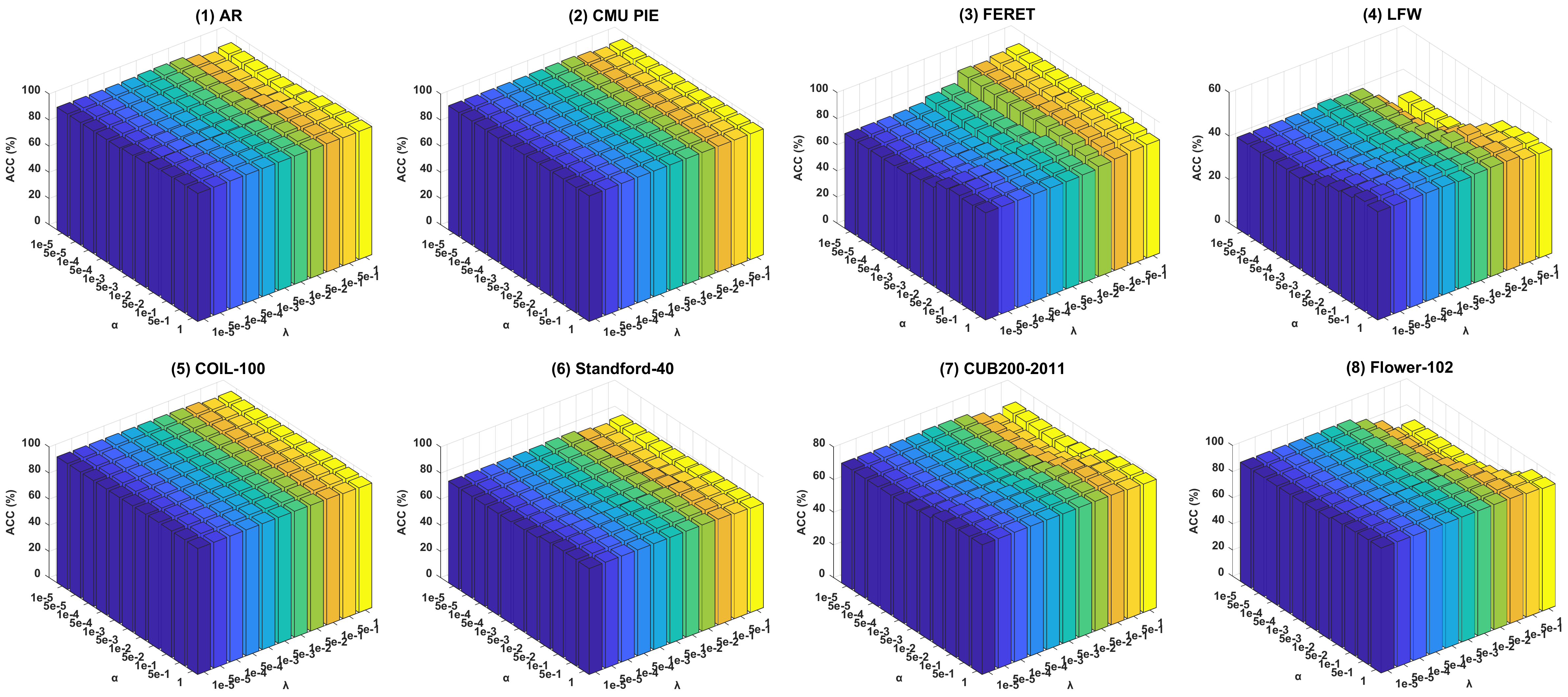}
\caption{The performance evaluation (\%) of FDLSR versus parameters $\alpha$ and $\lambda$ on eight datasets. Specifically, there are 14, 30, 5, 9, 30, 100, 30, and 20 training samples per class in the AR, CMU PIE, FERET, LFW, COIL100, Standford-40, CUB200-2011 and Flower-102 datasets, respectively. For each dataset, we randomly select one group of training and test data to conduct evaluation.  }
\end{figure*}

\subsection{Experiments in Large-Scale Image Recognition with Deep Features }
In this section, we report the results of the experiments designed to evaluate the performance of our FDLSR with VGG features comprehensively. Specifically, we compare FDLSR with state-of-the-art representation and LSR based classification algorithms on the following three challenging image datasets: the Stanford-40 Actions dataset \cite{yao2011human} for action recognition, the CaltechUCSD Birds (CUB200-2011) \cite{wah2011caltech} and Oxford 102 Flowers \cite{nilsback2008automated} datasets for fine-grained object recognition. 

\subsubsection{Experimental results on the Stanford-40 Actions dataset} The Stanford 40 Actions dataset consists of 9352 images of 40 human action classes and each action includes about 180-300 images. Following the sample grouping setting in \cite{xu2019sparse}\cite{yao2011human}, we randomly select 100 images per action as the training data and the remaining images are used for testing. We set the parameters $\alpha=1e-1$, $\beta=1e-2$, $\lambda=1e-2$ respectively in this dataset.

\subsubsection{Experimental results on the CUB200-2011 dataset} The Caltech-UCSD Birds (CUB200-2011) dataset includes 11788 bird images, widely used for fine-grained image recognition. There are 200 bird subjects in total and 60 images per subject. The main challenge for image classification is the variation in illumination, pose and viewpoint. In this dataset, the parameters are respectively set as $\alpha=1e-1$, $\beta=1e-2$, $\lambda=1e-1$. We use the publicly available training-testing grouping setting \cite{cai2016probabilistic}\cite{wah2011caltech}, in which nearly half of the images per subject are used as the training data and the other half of images are treated as the test data. 

\subsubsection{Experimental results on the Oxford 102 Flowers dataset}
The Oxford 102 Flowers dataset is also a fine-grained object image dataset which consists of 8189 images of 102 flower classes. The flowers appear at different scales, pose, and lighting conditions. Because of the relatively large intra-class variations but small inter-class variations of images, this dataset is very challenging for image classificaiton. We set the parameters as $\alpha=1e0$, $\beta=1e-2$, $\lambda=1e-3$ respectively in this dataset.

For these three datasets, we use the CNN features extracted by the VGG-verydeep-19 \cite{simonyan2014very} network which is pretrained using the ImageNet dataset \cite{deng2009imagenet}. All the data files of Matlab are provided by the authors of paper \cite{xu2019sparse}. In addition, we also compare our FDLSR employing deep features with two state-of-the-art deep learning methods, such as AlexNet \cite{krizhevsky2012imagenet} and VGG19 \cite{simonyan2014very} networks. The experimental results on these three datasets are listed in TABLE VI (some of the results have been taken from the original papers \cite{cai2016probabilistic} \cite{xu2019sparse}). The experimental results indicate that, with deep CNN features, the proposed FDLSR delivers better or at least comparable classification performance than state-of-the-art algorithms on different large-scale visual classification datasets. It proves to be an effective image classification algorithm.

\begin{figure*}[!t]
\centering
\includegraphics[scale=0.33]{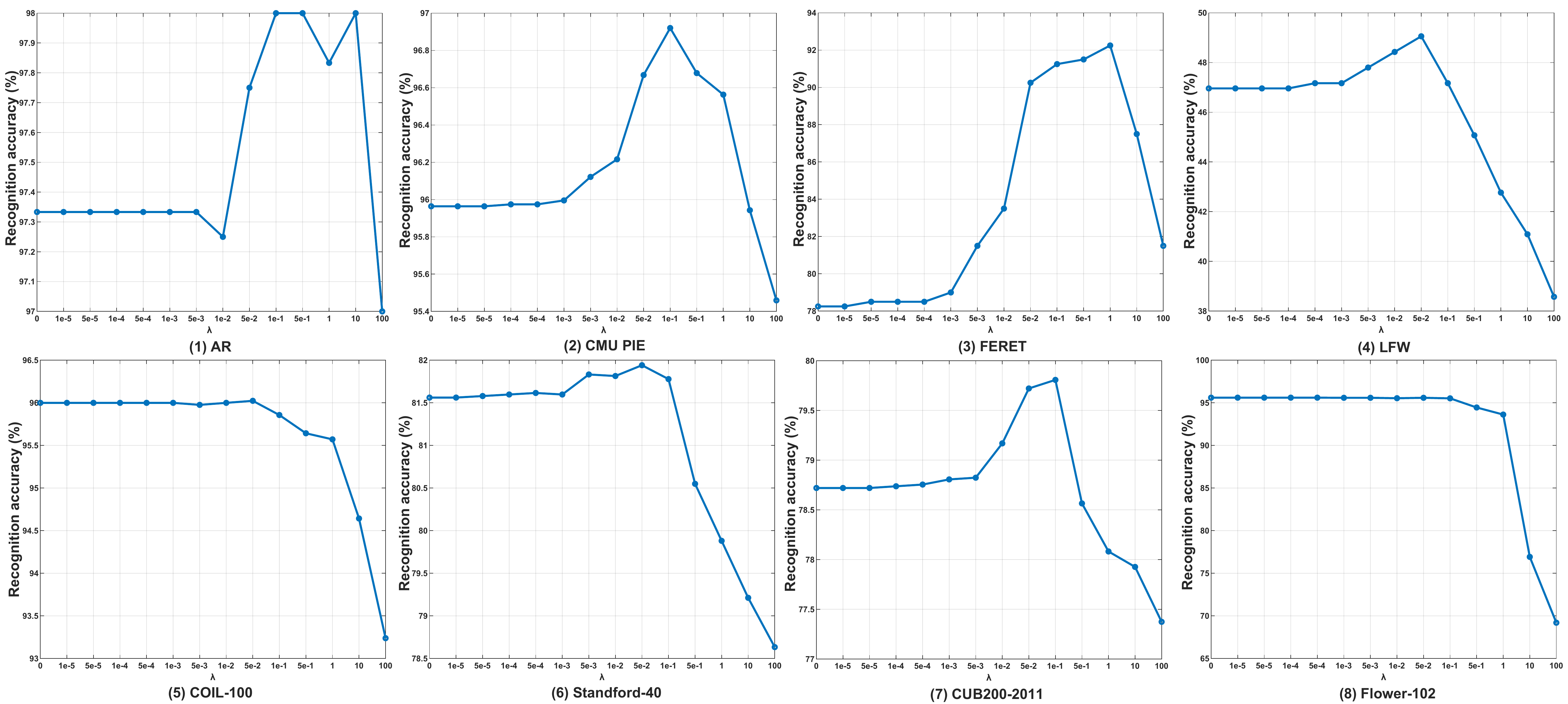}
\caption{The performance evaluation (\%) of FDLSR versus parameter $\lambda$ on eight datasets. The sample selection way is as the same as Fig. 3.}
\end{figure*}

\subsection{Parameter Sensitivity Analysis}
In this section, we carry out some experiments to analyze the parameter sensitivity of our FDLSR algorithm on  eight different datasets. For each dataset, we use a random split of training and test data to conduct validation. There are three parameters, $\alpha$, $\beta$ and $\gamma$, that need to be prudentlly tuned in our algorithm. Specifically, $\alpha$ and $\beta$ are used to weight the relaxation label learning term and the over-fitting avoiding term, respectively. $\lambda$ is used to balance the Fisher discrimination constraint. As the previous experiments show that the optimal value of $\beta$ did not change much from one dataset to another, we set $\beta$ to $1e-2$ for simplicity in the sensitivity analysis experiment and focus on discussing the sensitivity of parameters $\alpha$ and $\lambda$ while they assume values from a predefined set $\{1e-5, 5e-5,...,1e-1, 5e-1, 1\}$. The recognition results versus the values of parameters on eight datasets are shown in Fig. 3.  Relatively speaking, the performance of our FDLSR algorithm is not very sensitive to the values of parameters $\alpha$ and $\lambda$,  but the best recognition results are always achieved with large $\alpha$ and $\lambda$. In addition, as the main contribution of this paper is the incorporation of the Fisher discrimination criterion, we fix the parameters $\alpha$ and $\beta$ at their optimal values determined by cross-validation, and then study the effect of Fisher parameter $\lambda$ on the classification performance. As shown in Fig. 4,  the recognition accuracy of FDLSR gradually increases to its peak as the value of $\lambda$ falls close to $1e-1$, which proves that the Fisher criterion is indeed conducive to learning discriminative projection. Actually, when $\lambda=0$, our FDLSR degrades into the DLSR algorithm. Thus,  above results also prove the  statement that the Fisher criterion can be seen as a  complement of $\epsilon$-dragging method in the LSR model.

\section{Conclusion}
In this paper, we propose a novel Fisher discriminative least squares regression algorithm (FDLSR) for multi-class image recognition. FDLSR is based on  the framework of DLSR that uses the $\epsilon$-dragging technique to relax the strict binary labels. FDLSR is the result of a first-ever attempt to integrate the Fisher discrimination criterion and $\epsilon$-dragging technique into a unified LSR model. In FDLSR, these two constraints provide complementary information for learning the discriminative projection. Specifically, the $\epsilon$-dragging method is used to relax the original 'zero-one' labels and increase the margins between different classes. The Fisher criterion is applied to improve the intra-class similarity and compactness of the relaxed labels further to increase the class margins step by step during iteration. In this way, the learned regression labels are ensured to be not only relaxed but also sufficiently discriminative, and therefore producing efficient projection for classification. The experimental results demonstrate that the innovative combination leads to improved performance on several datasets, which is superior to that achieved by the state-of-the-art algorithms.

\section*{Acknowledgment}

This work was supported by the National Key Research and Development Program of China (Grant No. 2017YFC1601800), the National Natural Science Foundation of China (61672265, U1836218), the 111 Project of Ministry of Education of China (B12018), and the UK EPSRC (EP/N007743/1, MURI/EPSRC/DSTL, EP/R018456/1).

{\small
\bibliographystyle{ieee}
\bibliography{bare_jrnl}
}

\begin{IEEEbiography}[{\includegraphics[width=1in,height=1.25in,clip,keepaspectratio]{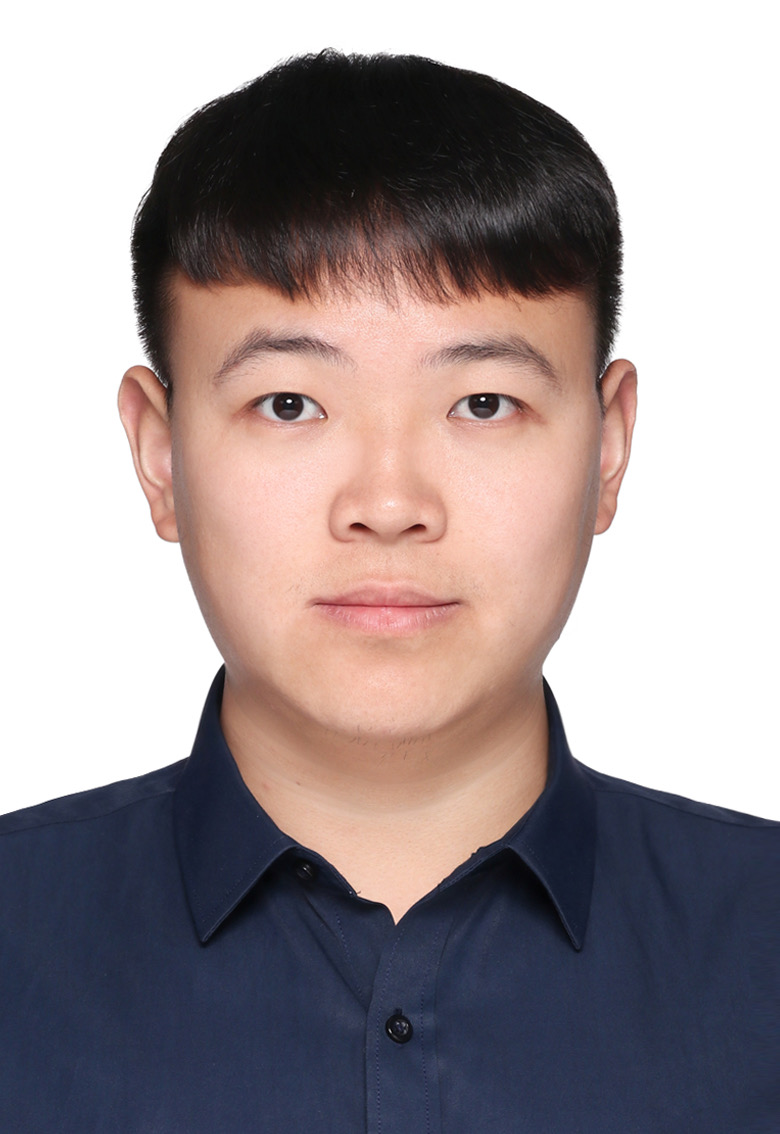}}]{Zhe Chen}
received the B.S. degree in the department of computer science and technology
from Hefei University, Hefei, China, in 2014, and the M.S. degree from the Jiangnan University,
Wuxi, in 2018. Currently, he is a PhD candidate in School of IoT Engineering, Jiangnan
University, Wuxi, China. His research interests include image recognition, dictionary learning
and sparse low-rank representation.
\end{IEEEbiography}

\begin{IEEEbiography}[{\includegraphics[width=1in,height=1.25in,clip,keepaspectratio]{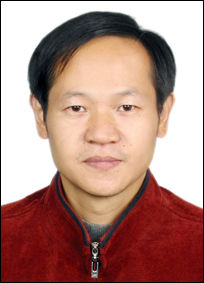}}]{Xiao-Jun Wu}
received the B.Sc. degree in mathematics from Nanjing Normal University, Nanjing, China, in 1991, and the M.S. degree and Ph.D. degree in pattern recognition and intelligent system from the Nanjing University of Science and Technology, Nanjing, in 1996 and 2002, respectively. From 1996 to 2006, he taught at the School of Electronics and Information, Jiangsu University of Science and Technology, where he was promoted to Professor. He has been with the School of Information Engineering, Jiangnan University, since 2006, where he is a Professor of Computer Science and Technology. He was a Visiting
Researcher with the Centre for Vision, Speech, and Signal Processing, University of Surrey, U.K., from 2003 to 2004. He has published over 200 papers. His current research interests include pattern recognition, computer vision, and
computational intelligence. He was a Fellow of the International Institute for Software Technology, United Nations University, from 1999 to 2000. He was a recipient of the Most Outstanding Postgraduate Award from the Nanjing
University of Science and Technology.
\end{IEEEbiography}

\begin{IEEEbiography}[{\includegraphics[width=1in,height=1.25in,clip,keepaspectratio]{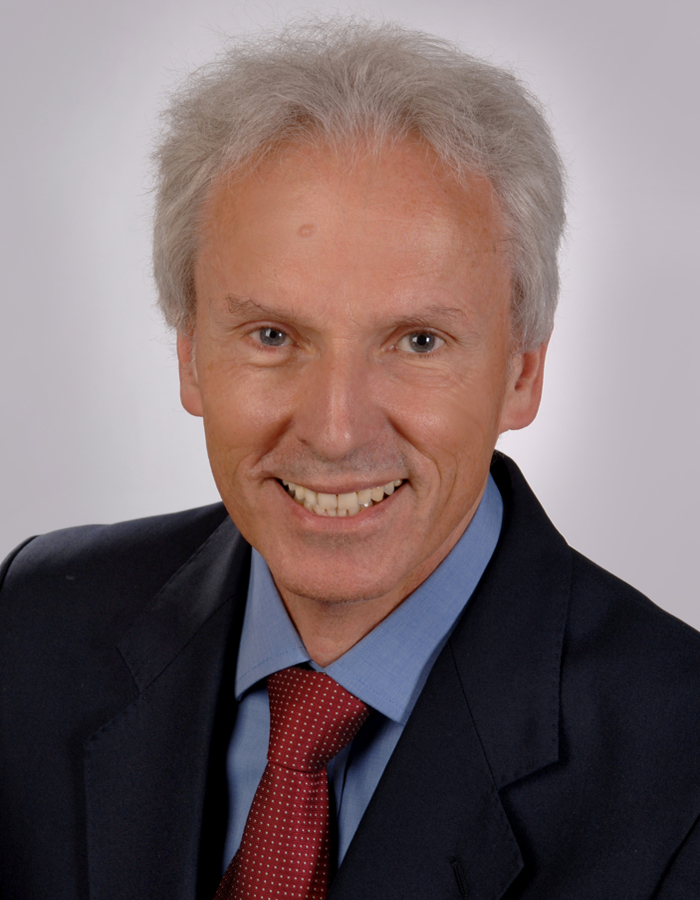}}]{Josef Kittler}
(M’74-LM’12) received the B.A., Ph.D., and D.Sc. degrees from the University of Cambridge, in 1971, 1974, and 1991, respectively. He is a distinguished Professor of Machine Intelligence at the Centre for Vision, Speech and Signal
Processing, University of Surrey, Guildford, U.K. He conducts research in biometrics, video and image
database retrieval, medical image analysis, and cognitive vision. He published the textbook Pattern Recognition: A Statistical Approach and over 700 scientific papers. His publications have been cited more than 60,000 times (Google Scholar).He is series editor of Springer Lecture Notes on Computer Science. He served on the Governing Board of the International Association for Pattern Recognition (IAPR) as one of the two British representatives during the period 1982-2005, President of the IAPR during 1994-1996.
\end{IEEEbiography}
\ifCLASSOPTIONcaptionsoff
  \newpage
\fi

\end{document}